\documentclass[conference]{IEEEtran}
\IEEEoverridecommandlockouts
\usepackage{cite}
\usepackage{amsmath,amssymb,amsfonts}
\usepackage{algorithmic}
\usepackage{graphicx}
\usepackage{textcomp}
\usepackage{xcolor}
\usepackage{booktabs}
\usepackage{multirow}
\usepackage{array}
\usepackage{pifont}
\usepackage{url}
\usepackage{hyperref}
\usepackage{balance}

\newcommand{\cmark}{\ding{51}} 
\newcommand{\xmark}{\ding{55}} 
\newcommand{\tmark}{\(\triangle\)} 

\def\BibTeX{{\rm B\kern-.05em{\sc i\kern-.025em b}\kern-.08em
    T\kern-.1667em\lower.7ex\hbox{E}\kern-.125emX}}

\begin{document}

\title{A Practical Synthesis of Detecting AI-Generated Textual, Visual, and Audio Content\thanks{This work was carried out with the support of CSPaper (\url{https://cspaper.org}), with contributions from Kai Xie, Lei You, Weiping Ding, Yong Du, Sven Salmonsson, Yumin Zhou, and Vilhelm von Ehrenheim.}}

\author{\IEEEauthorblockN{Lele Cao}
\IEEEauthorblockA{\textit{AI Labs, King/Microsoft} \\
\textit{R\&D, CSPaper}\\
Stockholm, Sweden \\
\url{https://orcid.org/0000-0002-5680-9031}}
}

\maketitle

\begin{figure}[t]
    \centering
    \includegraphics[width=\columnwidth]{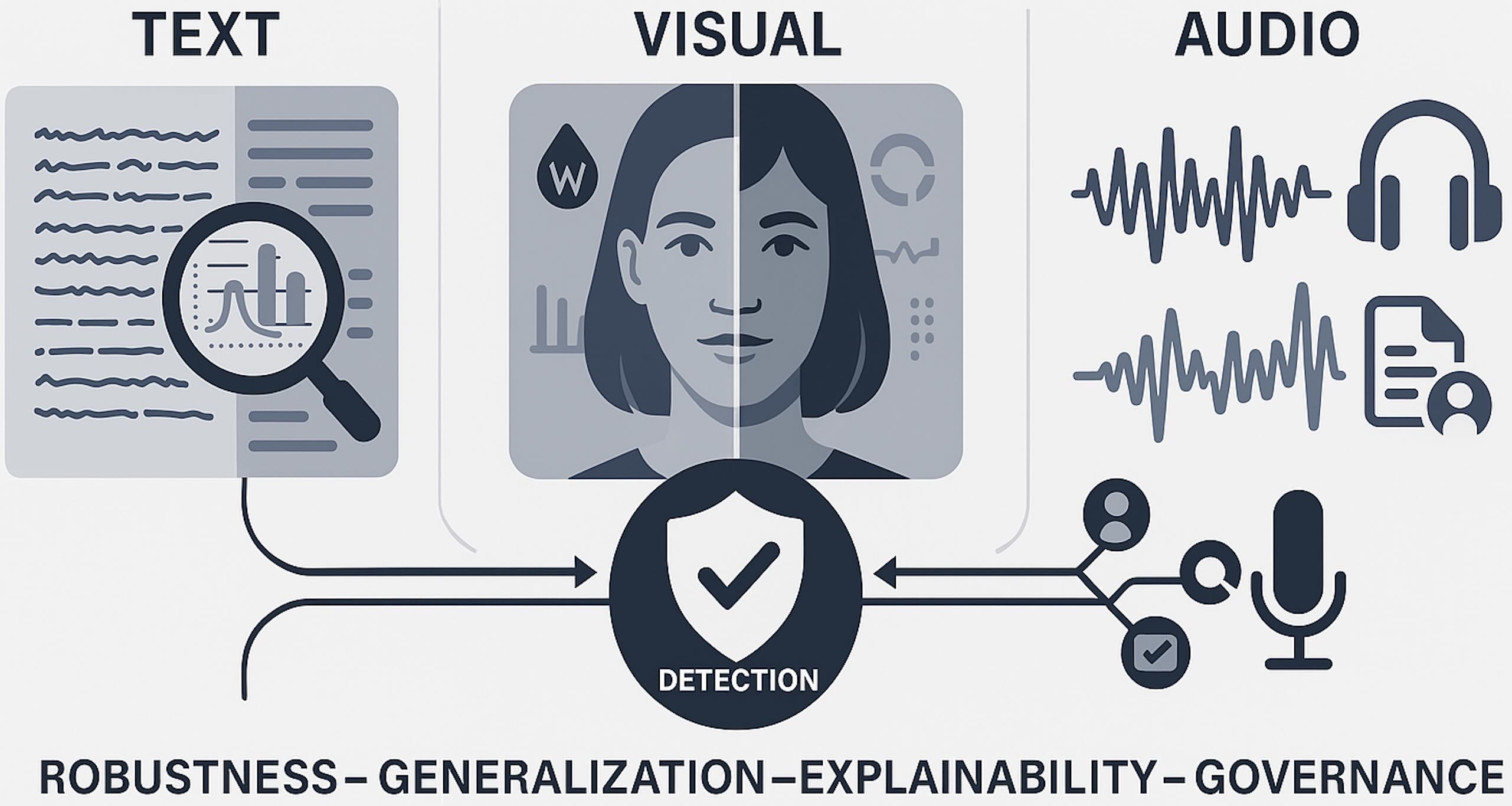}
    \label{fig:teaser}
    \vspace{-20pt}
\end{figure}

\begin{abstract}
Advances in AI-generated content have led to wide adoption of large language models, diffusion-based visual generators, and synthetic audio tools. However, these developments raise critical concerns about misinformation, copyright infringement, security threats, and the erosion of public trust. In this paper, we explore an extensive range of methods designed to detect and mitigate AI-generated textual, visual, and audio content. We begin by discussing motivations and potential impacts associated with AI-based content generation, including real-world risks and ethical dilemmas. We then outline detection techniques spanning observation-based strategies, linguistic and statistical analysis, model-based pipelines, watermarking and fingerprinting, as well as emergent ensemble approaches. We also present new perspectives on robustness, adaptation to rapidly improving generative architectures, and the critical role of human-in-the-loop verification. By surveying state-of-the-art research and highlighting case studies in academic, journalistic, legal, and industrial contexts, this paper aims to inform robust solutions and policymaking. We conclude by discussing open challenges, including adversarial transformations, domain generalization, and ethical concerns, thereby offering a holistic guide for researchers, practitioners, and regulators to preserve content authenticity in the face of increasingly sophisticated AI-generated media.
\end{abstract}


\section{Introduction}
\label{sec:intro}

Generative AI technologies have drastically changed how digital content is created, consumed, and transmitted worldwide \cite{chaka2024reviewing, li2023origin}. Large language models (LLMs) now consistently produce text that emulates human writing style \cite{wu2024survey, tang2024science}, diffusion and GAN (Generative Adversarial Networks) based frameworks yield photorealistic images and videos \cite{ho2020denoising, borji2023qualitative}, and advanced text-to-speech (TTS) systems synthesize voices that rival real speakers \cite{wang2020asvspoof, li2024audio}. While these breakthroughs offer profound benefits to creative industries, data augmentation, and accessibility solutions, they also open the floodgates for significant societal challenges.

Unlike previous surveys that have largely focused on single modalities or isolated detection strategies, our work uniquely integrates detection approaches across textual, visual, and audio domains. We provide a comprehensive analysis of state-of-the-art techniques while also highlighting emerging trends such as watermarking, self-supervised learning, generative model fingerprinting, and human-in-the-loop verification. This holistic perspective, reinforced by practical case studies, differentiates our survey and underscores its relevance for both research and industrial applications.

Maliciously generated or manipulated content can propagate disinformation, orchestrate social engineering attacks, breach security protocols, undermine academic integrity, and violate intellectual property rights \cite{valiaiev2024detection, crothers2023machine}. From deceptively plausible news articles to convincing audio recordings impersonating public figures, adversaries increasingly create forgeries that elude basic detection.

Under increasing global concerns, the quest for robust detection of AI-generated textual, visual and audio content has intensified \cite{yang2023survey, wu2024survey}. Governments, academic institutions, technology platforms, and industries alike seek solutions to discern authentic content from synthetic. Addressing this need has led to the proliferation of detection strategies, ranging from visual artifact analysis to linguistic fingerprinting, watermark verification, and classification approaches.

The paper is organized as follows. Section~\ref{sec:background} outlines the conceptual foundations of AI-generated content, including key generation approaches and threat models. Section~\ref{sec:text} focuses on text-based detection strategies, from prompting-based classification to watermarking-based schemes. Section~\ref{sec:visual} covers visual detection methods for AI-generated images and videos, emphasizing artifact analysis, watermarking, and advanced video forensics. Section~\ref{sec:audio} addresses synthetic audio detection, highlighting speech deepfake risks and music plagiarism. Section~\ref{sec:case} provides real-world case studies in academia, journalism, and creative industries. Section~\ref{sec:discussion} surveys cross-modal insights, challenges, and future directions, including emerging self-supervised and generative model fingerprinting techniques. Finally, we conclude by underscoring the need for a multi-stakeholder and adaptive approach to preserve digital authenticity in the rapidly evolving generative AI landscape.

\subsection*{Contributions and Unifying Taxonomy}
\noindent\textbf{What is new.} Beyond synthesizing prior work, we contribute:
\begin{itemize}
    \item \textit{A cross-modal taxonomy} unifying textual, visual, and audio detectors into five families: (i) \emph{signal/statistical} (stylometry, pixel/spectral cues), (ii) \emph{model-likelihood} (perplexity/curvature), (iii) \emph{supervised classifiers} (fine-tuned detectors), (iv) \emph{provenance mechanisms} (watermarking/fingerprinting and content credentials), and (v) \emph{retrieval- and consistency-based} (nearest-neighbor retrieval, cross-modal agreement checks).
    \item \textit{A threat model matrix} mapping attacker capabilities to defenses across modalities (Table~\ref{tab:threat-matrix}), clarifying realistic deployment choices and gaps.
    \item \textit{Actionable deployment notes} for retrieval-based text detection (mitigating paraphrasing), temporal/physiological video forensics, and real-time multilingual audio detection.
    \item \textit{Reproducible resources} linking datasets, leaderboards, and open-source implementations (Appendix~\ref{app:resources}).
\end{itemize}

\section{Background on AI-Generated Content}
\label{sec:background}

To contextualize the approaches for detection, we provide a brief overview of the primary mechanisms behind AI-generated content and typical threat models. 

\subsection{Generative Modeling Paradigms}

Three major paradigms underlie most modern AI-based content generation:

\begin{itemize}
    \item Autoregressive models: Exemplified by GPT-type architectures, these models predict tokens sequentially to generate coherent text, code snippets, lyrics, or even image patches (by ViT - Vision Transformer) \cite{tang2024science, bhattacharjee2024fighting}. Because large parameter counts are typically needed for strong performance, they are often referred to as LLMs in NLP (natural language processing) domain.
    \item GANs: Operate on the principle of a generator and discriminator in competition, widely used for producing images, style transfer, and deepfake videos \cite{borji2023qualitative}. 
    \item Diffusion models: Gradually denoise random noise into highly realistic images or audio. Known for generating crisp visuals (e.g.,~Stable Diffusion \cite{ho2020denoising} and DALL-E\footnote{\url{https://openai.com/index/dall-e-3}}) and increasingly refined speech and music \cite{li2024milling}.
\end{itemize}

These architectures have been refined to handle specialized tasks: style-based face generation, voice cloning, text-to-music composition, etc. They can also be composited, e.g., LLM-driven prompts controlling diffusion image generation, to produce multi-modal AI outputs.

\subsection{Threat Models and Malicious Use Cases}

\textbf{Disinformation campaigns} involve automated text production at scale, which can flood social media with politically motivated messages, conspiracy theories, or phony press releases, effectively shaping public opinion \cite{crothers2023machine, li2023origin}. \textbf{Identity theft and fraud} can be facilitated by synthetic voice or video mimicking a corporate executive or family member, enabling financial scams, social engineering, or blackmail \cite{wang2020asvspoof, yi2023audio}. \textbf{Academic dishonesty} arises when students or researchers generate entire essays, lab reports, or dissertations using LLMs \cite{valiaiev2024detection}, undermining educational integrity. \textbf{Plagiarism and copyright infringement} may occur as AI composition tools create music or artwork that partially copies existing works or inadvertently violates intellectual property, posing a serious challenge \cite{li2024aigmdetection, li2024detectingmachine}. \textbf{Manipulation in e-commerce or branding} is also a concern, as product images or videos might be faked to damage a competitor’s brand or mislead customers about product quality \cite{lin2024detecting}.

The overall challenge is that as AI-based generation quality improves, naive or conventional authenticity checks fail. In response, a vibrant research ecosystem focuses on robust detection measures, as discussed next.

\begin{table*}[t]
\centering
\caption{Attacker capabilities vs.\ defenses across modalities (text / image-video / audio). \cmark\ likely effective; \tmark\ partially effective or context-dependent; \xmark\ ineffective. ``Prov.'' includes watermarking/fingerprinting and provenance standards (e.g., C2PA).}
\label{tab:threat-matrix}
\begin{tabular}{p{4.2cm}ccc|ccc|ccc}
\toprule
\multirow{2}{*}{\textbf{Attacker capability}} & \multicolumn{3}{c|}{\textbf{Text defenses}} & \multicolumn{3}{c|}{\textbf{Image/Video defenses}} & \multicolumn{3}{c}{\textbf{Audio defenses}} \\
 & Stat./LLM & Sup./Ensemble & Prov. & Pixel/Freq. & Temporal & Prov. & Pipeline/NN & End-to-end & Prov.\\
\midrule
Zero-effort (raw model output) & \cmark & \cmark & \cmark & \cmark & \cmark & \cmark & \cmark & \cmark & \cmark \\
Paraphrasing / style transfer & \tmark & \tmark & \tmark$^{*}$ & \tmark & \tmark & \tmark & \tmark & \tmark & \tmark \\
Heavy compression / resampling & \tmark & \tmark & \cmark & \tmark & \tmark & \cmark & \tmark & \tmark & \cmark \\
Cropping / rotation / time-stretch & \tmark & \tmark & \cmark & \tmark & \tmark & \tmark & \tmark & \tmark & \tmark \\
Model-switching (unmarked models) & \tmark & \tmark & \xmark & \tmark & \tmark & \xmark & \tmark & \tmark & \xmark \\
Adversarial perturbations & \tmark & \tmark & \tmark & \tmark & \tmark & \tmark & \tmark & \tmark & \tmark \\
Cross-lingual / domain shift & \tmark & \tmark & \tmark & \tmark & \tmark & \tmark & \tmark & \tmark & \tmark \\
Splicing / synthetic-real mixing & \tmark & \tmark & \tmark & \tmark & \tmark & \tmark & \tmark & \tmark & \tmark \\
\bottomrule
\end{tabular}

\begin{minipage}{0.98\linewidth}
\vspace{5pt}
\scriptsize $^{*}$ Several works show that paraphrasing can substantially degrade watermark detection, though more robust schemes continue to be proposed \cite{kirchenbauer2023watermark, kuditipudi2023robust, hou2023semstamp, krishna2023paraphrasing, zhao2024provable, pruthi2024revisiting}. Provenance standards like C2PA provide complementary, opt-in authenticity metadata \cite{c2paSite}.
\end{minipage}
\end{table*}

\section{Detecting AI-Generated Text}
\label{sec:text}

\subsection{Key Motivations and Constraints in Textual Detection}

The written word remains a cornerstone of communication in academia, journalism, business, and everyday conversations. LLM-generated content can be harnessed productively, but the ease of generating vast, coherent text also creates new vulnerabilities, including misinformation spread, spam, impersonation, and academic plagiarism~\cite{chaka2024reviewing, wu2024survey}.

Several major constraints hinder effective textual detection. \textbf{Language diversity} poses a significant challenge, as tools must function across different languages, dialects, and text domains (technical, legal, creative). \textbf{Model evolution} further complicates detection -- systems trained on older text generators such as GPT-2~\cite{radford2019language} often underperform when confronted with more advanced models like GPT-4~\cite{achiam2023gpt} or Bard~\cite{nyberg2022bard}. Additionally, \textbf{paraphrasing attacks} exploit simple rephrasings to mask many stylistic signals of machine-generated text~\cite{krishna2024paraphrasing}.

\subsection{Approaches}

\subsubsection{LLM prompting and zero-shot methods}
One popular approach leverages an external LLM to classify whether a piece of text is AI-generated or not \cite{bhattacharjee2024fighting}. With carefully crafted prompts, these zero-shot methods can achieve moderate accuracy quickly. However, sensitivity to prompt design and the adversarial gap between the LLM detector and the text generator are common issues \cite{taguchi2024impact}. Recent zero-shot probabilistic curvature methods (e.g.,~DetectGPT and Fast-DetectGPT) trade accuracy and compute in different ways \cite{mitchell2023detectgpt, bao2023fastdetectgpt}.

\subsubsection{Linguistic and statistical signatures}
Traditional stylometric features (e.g., function words, syntax complexity, average phrase length) have long been used in authorship attribution \cite{mitchell2023detectgpt, hans2024spotting}. More modern detection focuses on computing perplexity or log-likelihood using reference language models, observing that LLM-generated text tends to show distinctive probability distributions. Additionally, specialized white-box methods can measure rank ordering of tokens if the generating model is partially known \cite{su2023detectllm}.

\subsubsection{Supervised classification (training-based)}
Labeled corpora of AI versus human text enable fine-tuning of large pre-trained transformers like RoBERTa or T5 to discriminate synthetic text \cite{liu2019roberta, chen2023gpt}. Researchers improve robustness with adversarial training sets that contain paraphrased or AI-generated passages shifted in style. 
Tools like GPTZero \cite{tian2024multiscale} and RADAR \cite{hu2023radar} exemplify advanced supervised detectors. However, assembling high-quality, representative training data remains a challenge, especially as new generator architectures emerge frequently.

\subsubsection{Watermarking for AI text}
Cooperative watermarking modifies text generation at token selection time, embedding an imperceptible pattern in the distribution of words or punctuation \cite{kirchenbauer2023watermark, sadasivan2023can}. A verifier can detect such patterns after the fact. While promising for major industrial LLMs that adopt the standard, watermarking fails if malicious or open-source models do not embed it, or if paraphrasing disrupts the signal \cite{hou2023semstamp, kuditipudi2023robust, pruthi2024revisiting, zhao2024provable}. Complementary provenance standards (e.g., C2PA) aim to capture capture-chain information when available \cite{c2paSite}.

\subsubsection{Ensemble and multi-feature systems}
To mitigate single-method vulnerabilities, some frameworks combine perplexity-based signals, style analysis, embedding-based classification, and watermark checks \cite{ong2023applying, zeng2024detecting}. By fusing different perspectives, these ensembles often achieve higher accuracy. The trade-off is system complexity and the need for sufficiently large training resources.

\subsubsection{Retrieval-augmented defenses against paraphrasing (deployment notes)}
Paraphrasing can significantly degrade both watermark-based and likelihood-based detectors, yet retrieval proves effective: by searching for near-duplicates of candidate passages in a large corpus of model outputs or web-scale text, paraphrase-derived copies can be surfaced \cite{krishna2024paraphrasing}. In practice we recommend: (i) maintain an index of recent LLM outputs or prompt-completion pairs; (ii) use character $n$-gram shingles (e.g., 5$\sim$13) and embedding-based ANN search; (iii) combine retrieval scores with a calibrated detector score (e.g., logistic fusion); (iv) perform chunked matching (200 to 400 tokens) with overlap to handle local paraphrases; (v) log decisions for human audit in high-stakes settings (academia/journalism). Fast zero-shot curvature detectors \cite{bao2023fastdetectgpt, mitchell2023detectgpt} can act as efficient first-stage filters before expensive retrieval.

\section{Detecting AI-Generated Visual Content}
\label{sec:visual}

\subsection{Motivations and Real-World Impact}

AI-generated images and videos, often created via GANs or diffusion models, enable powerful visual illusions \cite{borji2023qualitative, lin2024detecting}. Notable concerns include \textbf{political misinformation}, where fabricated news images depict fictional events; \textbf{financial fraud}, involving misleading product visuals or manipulation of stock markets; \textbf{harassment and defamation}, as seen in deepfake pornography or face swaps designed to humiliate victims; and \textbf{intellectual property theft}, such as art or design forgery that undermines artists’ livelihoods.

\subsection{Detection Methodologies}

\subsubsection{Observation and manual inspection}
Human experts can sometimes detect unnatural artifacts in lighting, shadows, perspective, or anatomical features \cite{kamali2024distinguish}. Context-based checks (e.g., unrealistic historical detail) also help. However, manual inspection is subjective, time-consuming, and not scalable for large volumes of online images.

\subsubsection{Model-based artifact analysis}
Algorithms analyze pixel-level statistics or frequency-domain features. For instance, Fourier transform reveals periodic textures characteristic of upsampling procedures in GANs or consistent small-scale noise from diffusion \cite{mavali2024fake, saberi2024robustness}. White-box strategies exploit knowledge of the generator's pipeline (e.g., measuring the likelihood under the reverse diffusion process).

\subsubsection{Black-box deep learning classifiers}
With large amount of labeled real/fake data, CNN (convolutional neural network) based classifiers (ResNet, EfficientNet) learn discriminative cues \cite{saskoro2024detection}. Ensemble approaches combine multiple model outputs or domain-specific sub-networks (e.g., focusing on faces vs. backgrounds) to bolster accuracy.

\subsubsection{Watermarking for visual media}
When the generative pipeline is compliant, watermarks are embedded. These vary from invisible spatial pixel encodings to frequency manipulations \cite{kirchenbauer2023watermark, liu2024semantic}. However, simple image transformations, such as crop or rotation, can weaken naive watermarks \cite{hou2023semstamp, sharma2024review} unless specifically designed for robustness. Content credentials such as C2PA can attach tamper-evident provenance metadata on capture and edit chains \cite{c2paSite}.

\subsubsection{Temporal/physiological forensics for video}
Beyond per-frame artifacts, temporal and bio-signal cues are highly informative. Methods exploit phoneme-viseme inconsistencies in lip motion \cite{agarwal2020phoneme}, remote photoplethysmography (rPPG) from subtle skin color changes \cite{ciftci2019fakecatcher}, eye-blink and micro-expression patterns \cite{li2018blink}, optical flow coherence and head-pose/motion geometry \cite{nguyen2024videofact, bai2024ai, chang2024what, vahdati2024beyond}. These cues are often more resilient to spatial filtering but can be sensitive to heavy compression and low frame rates.

\subsection{Datasets and Benchmarks}

Well-known datasets include CelebA-HQ~\cite{karras2018progressive} for facial images and LAION~\cite{schuhmann2022laion} for broad image domains. Large-scale synthetic benchmarks like GenImage (1M+ pairs; diverse generators and degradations) offer cross-generator and degraded-image evaluation \cite{zhu2023genimage, genimageNeurIPS23}. Community-maintained leaderboards and consolidated testbeds (e.g., AIGCDetect Benchmark) enable reproducible comparisons and track generalization to newer models \cite{aigcdBench, aigcdPages}. For video, FaceForensics++~\cite{rossler2019faceforensics++}, DFDC~\cite{dolhansky2020deepfake}, and Celeb-DF are commonly used \cite{tiwari2024ai}. However, frequent advancement of the generative model requires continuous expansion of the detection dataset.

\subsection{Limitations and Outlook}

\textbf{Compression sensitivity} is a major challenge, as downsampling, scaling, or re-encoding can obscure forensic traces~\cite{saberi2024robustness, yan2024sanity}. \textbf{Generalization to new architectures} remains difficult since tools often lag behind novel generator types, such as next-generation diffusion or hybrid models. \textbf{Ethical implications} also complicate detection efforts: large-scale scanning of user images for potential deepfakes raises privacy concerns, while overly aggressive detection can flag benign content as suspicious, ultimately hurting user trust.

\section{Detecting AI-Generated Audio Content}
\label{sec:audio}

\subsection{Risks and Use Cases}

Synthetic audio has rapidly evolved thanks to neural TTS, voice conversion (VC), and audio diffusion. High-fidelity speech generation from minimal samples can facilitate voice impersonation~\cite{wang2020asvspoof, yi2023audio} or mislead detection systems in telephony security. Meanwhile, AI-composed music raises legal problems on originality, licensing, and plagiarism~\cite{li2024milling}.

Use cases of audio detection span several domains. \textbf{Voice deepfake forensics} is essential in contexts such as law enforcement, banking, or enterprise authentication systems, where verifying speaker identity is critical~\cite{muller2022does, hussain2022forensic}. \textbf{Music authenticity} matters for streaming services or record labels that aim to detect GenAI music to safeguard artists' rights~\cite{li2024detectingmachine}. \textbf{Real-time moderation} is crucial for conference platforms that filter suspicious speech to prevent social engineering attacks.

\subsection{Detection Techniques}

\subsubsection{Pipeline classifiers}
Features such as Mel frequency cepstral coefficients (MFCC), Linear frequency cepstral coefficients (LFCC) spectrogram-based descriptors are extracted and then fed into machine learning models (SVM, XGBoost, or CNN) \cite{yi2023audio, li2024audio}. They typically rely on analyzing subtle cues in pitch, timbre, and vocal fold dynamics.

\subsubsection{End-to-end neural approaches}
Powerful audio deepfake detectors ingest raw waveforms or full spectrograms using deep architectures like SincNet, Wav2Vec2.0~\cite{baevski2020wav2vec}, or CRNN \cite{ravanelli2018speaker, pascu2024generalizable}. These systems can learn complex patterns indicative of synthetic audio, such as unnatural transitions or missing microprosody. However, performance can degrade with domain shifts (e.g., new TTS pipelines, different languages) or background noise.

\subsubsection{Music detection specifics}
Music detection often analyzes melodic structure, chord progressions, or repetitive patterns \cite{li2024milling, rao2022melody}. Large neural networks trained on real vs. AI-generated music can spot overly mechanical or simplistic progressions \cite{li2024aigmdetection, comanducci2024fakemusiccaps}.

\subsubsection{Audio watermarking}
Similar to the textual and visual domain, watermarking can be inserted into the synthetic audio during generation, using imperceptible frequency modulations or phase shifts \cite{deepmind2024synthid, liu2024semantic}. The watermark reveals the audio's origin; however, many transformation (tempo shift, reverb, denoising) can weaken naive watermarking signals.

\subsection{Operational considerations: latency and multilingual robustness}
\textbf{Streaming/real-time}: For telephony fraud prevention and live moderation, detectors must operate with very low latency. Sub-second chunking (e.g., 0.5--1.0\,s hops with 2--3\,s context windows) enables timely alerts, while causal CNN or Conformer backbones with look-ahead $\le$200\,ms help maintain responsiveness. Incorporating on-device VAD (voice activity detection) further reduces computational overhead and lowers false-alarm rates. \textbf{Calibration}: Detector performance should be tuned to in-domain EERs (equal error rates). Dynamic thresholding strategies that adapt to input SNR levels can significantly improve stability in noisy or variable conditions. \textbf{Multilingual/broad-accent}: Results from the ASVspoof evaluations, which include the \emph{Logical Access (LA)} track for algorithmically generated speech, the \emph{Physical Access (PA)} track for replayed audio, and the \emph{Deepfake (DF)} track for highly realistic neural synthesis, consistently show sharp performance drops under cross-corpus and cross-condition settings. These findings underscore the importance of training with diverse languages and accents, followed by periodic domain adaptation \cite{asvspoof2019, asvspoof2021sum, asvspoof2021site}. Robustness can be further enhanced with self-supervised pretraining methods (e.g., HuBERT, Wav2Vec2) and test-time augmentations such as noise injection or codec simulation.

\section{Case Studies in Practice}
\label{sec:case}

To illustrate how detection strategies apply in different domains, we highlight several real-world contexts where AI-generated content is already a pressing concern.

\subsection{Academic Integrity and Higher Education}

Universities face surging usage of LLM tools for assignments and research papers. In many cases, naive plagiarism checks fail to detect newly generated text. Some institutions adopt specialized systems (e.g., GPTZero\footnote{\url{https://gptzero.me}}, Turnitin's AI detection\footnote{\url{https://www.turnitin.com}}) that combine perplexity measures, stylometric analysis, and partial reference matching \cite{wu2024survey, valiaiev2024detection}. However, concerns about privacy (scanning entire student submissions) and false positives remain. Many universities are establishing policies requiring students to label or disclaim AI assistance.

\subsection{Newsrooms and Journalistic Fact-Checking}

Misinformation campaigns are increasingly complex due to auto-generated text, manipulated images (e.g., fabricated protest scenes), and deepfake videos of political leaders. Journalists use hybrid detection pipelines: a first-level automated classifier flags suspicious content, which is then reviewed by human fact-checkers \cite{crothers2023machine}. They also rely on watermark or metadata checks if major AI model providers tag their outputs. Major media platforms and social networks have begun integrating these systems into their content moderation workflows, combining automated screening with expert review to minimize false positives.

\subsection{Law Enforcement and Legal Proceedings}

Synthetic audio or video can compromise evidence authenticity. Forensic experts apply advanced image, audio forensics, and motion analysis tools to verify recordings \cite{tiwari2024ai, hussain2022forensic}. They may also cross-reference biometric cues, such as lip movement, voice biometrics, or EKG-like signals from speech waveforms. Courts increasingly grapple with how to interpret detection tool outputs, calling for transparent and explainable detection methods.

\subsection{Creative Industries and Content Platforms}

Content creators worry about plagiarism from AI tools that replicate their style or incorporate copyrighted material~\cite{li2024aigmdetection}. Music streaming platforms experiment with classifier-based scanning of newly uploaded tracks for suspicious patterns. Some are exploring watermark enforcement with partial success~\cite{deepmind2024synthid}. Visual artists are also advocating for improved detection of unauthorized use of their work, prompting platforms to implement hybrid approaches that combine automated and manual review processes.

For researchers interested in experimental validation and reproducibility, numerous open-source frameworks and benchmark leaderboards (e.g.,~DFDC, FaceForensics++, Audio Deepfake datasets) are available. We reference these resources to facilitate further exploration of state-of-the-art methods, and encourage readers to consult~\cite{cao2025practical} for a more practical and comprehensive guide.

\section{Cross-Modal Insights, Challenges, and Future Directions}
\label{sec:discussion}

\subsection{Unified Themes Across Modalities}

Despite modality differences, several consistent themes arise. 
\begin{enumerate}
    \item \textit{Arms race with generators}: As generative models advance, older detection strategies degrade, requiring frequent retraining or adaptation~\cite{tang2024science, mireshghallah2024smaller}.
    \item \textit{Vulnerability to perturbations}: Slight paraphrasing in text, minor image edits (e.g., crop and color shift), or audio pitch/time adjustments can bypass naive methods~\cite{krishna2024paraphrasing, saberi2024robustness}.
    \item \textit{Watermarking as partial solution}: Watermarking is useful if widely adopted by model providers, but it’s easily circumvented by uncooperative or malicious providers~\cite{kuditipudi2023robust, pruthi2024revisiting}.
    \item \textit{Ensemble or multimodal methods}: Combining multiple cues (statistical, watermark, contextual) and bridging text–image–audio modalities yields more robust detection~\cite{ong2023applying, zeng2024detecting}.
    \item \textit{Ethical pitfalls}: Systemic large-scale scanning can infringe on user privacy, and erroneous misclassifications can harm reputations.
\end{enumerate}

\subsection{Deeper analysis of watermark robustness}
Empirical and theoretical studies report mixed results on watermark robustness under paraphrase, editing, and translation. While Unigram/semantic variants and adaptive schemes seek provable or empirical robustness \cite{zhao2024provable, kuditipudi2023robust, hou2023semstamp}, other analyses demonstrate learnability or reverse-engineering of green-list partitions and sharp drops in detection rates under targeted paraphrasing \cite{pruthi2024revisiting, kirchenbauer2023reliability}. We therefore position watermarking as one ingredient in a broader provenance strategy alongside content credentials (C2PA) and out-of-band logging for cooperative platforms \cite{c2paSite}.

\subsection{Concrete pathways to address open challenges}
\textbf{Adversarial robustness}: Strengthening resilience requires transformation-consistent training (such as paraphrasing or back-translation for text, crop/resize/codec chains for images, and time-stretching or noise perturbations for audio) combined with adversarial example mining and confidence calibration through temperature scaling. \textbf{Domain generalization}: To cope with unseen generators and distribution shifts, promising strategies include generator-aware data augmentation, continual learning with replay of past samples, and retrieval-based regularizers that leverage nearest-neighbor consistency. \textbf{Explainability}: Enhancing interpretability involves highlighting salient tokens, regions, or frames and linking them to human-understandable cues, for example, viseme–phoneme alignment~\cite{agarwal2020phoneme} in video or rPPG signals in facial imagery. \textbf{Governance}: Beyond algorithmic techniques, standardized disclosures and verifiable audit trails, implemented via provenance metadata (e.g., C2PA manifests), should be paired with opt-in watermarking whenever feasible to ensure accountability and transparency.

\subsection{Dataset gaps and mitigation}
Current benchmarks are heavily skewed toward \emph{faces} and \emph{English} text, leaving major gaps in other domains. Non-face imagery (e.g., documents, user interfaces, medical scans, remote-sensing data) and low-resource languages remain underrepresented, limiting generalization to diverse real-world scenarios. In audio, multi-lingual, multi-accent, and codec-diverse corpora are still scarce despite notable progress from ASVspoof \cite{asvspoof2019, asvspoof2021sum}. To close these gaps, we recommend three complementary strategies: (i) targeted data collection in underrepresented modalities and languages; (ii) synthetic hard-negative generation to expose detectors to adversarially challenging examples; and (iii) cross-temporal benchmarks that explicitly test robustness against newly released generators, such as GenImage and evolving AIGCDetect testbeds \cite{zhu2023genimage, aigcdBench, genimageNeurIPS23}.

\section{Conclusion and Broader Reflections}

As AI generation systems become more sophisticated and widespread, detection is a critical fortress to uphold authenticity, trust, and accountability in the digital ecosystem. This paper has surveyed leading techniques for distinguishing AI-generated text, images, video, voice, and music, with attention to the motivations, current approaches, challenges, and ethical underpinnings across each domain. Our investigation underscores the following overarching lessons:

\begin{itemize}
    \item No silver bullet: Each detection category, text, visual, or audio, relies on complementary signals (statistical, watermark-based, manual observation), yet none are foolproof against adaptive adversaries.
    \item Continual adaptation: Generative models evolve quickly. Detectors must be regularly updated, often requiring new training data from newly released generation architectures.
    \item Watermarking potential and pitfalls: Watermarks show promise if major platforms adopt them systematically, but they are easily circumvented by malicious or open-source models.
    \item Contextual and human-AI collaboration: Real-world detection extends beyond pure algorithmic classification. Human oversight, context checks, retrieval-based cross-referencing, and specialized domain knowledge remain pivotal, especially in high-stakes use cases.
    \item Ethical complexity: Overly invasive or inaccurate detection can harm user privacy and trust, while under-detection fuels misinformation and fraud. Balancing these risks requires responsible governance.
\end{itemize}

Looking forward, deeper integration across modalities, advanced retrieval-based or self-supervised approaches, improved adversarial robustness, and generative model fingerprinting will define the next generation of AI-content detectors. Collaboration among academia, industry, policymakers, and civil society is paramount to develop globally recognized standards and frameworks ensuring that generative AI can flourish as a positive force, while preserving integrity and truthfulness in digital media.

For further learning, we suggest to explore open-source tools, benchmark datasets, and books like~\cite{cao2025practical} to accelerate progress in this fast-moving field.

\appendices
\section{Reproducibility Resources (Datasets, Detectors, Scripts)}
\label{app:resources}

To support reproducibility and practical experimentation, we summarize widely used datasets, detectors, and code resources across modalities in Table~\ref{tab:resources}. This table centralizes pointers that can serve as entry points for researchers and practitioners.

\begin{table}[h]
\centering
\caption{Key reproducibility resources for AI-generated content detection.}
\label{tab:resources}
\begin{tabular}{p{1.5cm} | p{6.3cm}}
\toprule
\textbf{Modality} & \textbf{Representative Resources} \\
\midrule
Text & DetectGPT, Fast-DetectGPT (zero-shot curvature) \cite{mitchell2023detectgpt, bao2023fastdetectgpt}; retrieval-based defenses \cite{krishna2024paraphrasing}; semantic/robust watermarks \cite{hou2023semstamp, kuditipudi2023robust, zhao2024provable} \\
\addlinespace
\midrule
Image\newline /Video & FaceForensics++ \cite{rossler2019faceforensics++}; DFDC \cite{dolhansky2020deepfake}; GenImage (1M+ pairs) \cite{zhu2023genimage, genimageNeurIPS23}; community leaderboards \cite{aigcdBench, aigcdPages}; temporal/physiological cues such as viseme mismatch \cite{agarwal2020phoneme}, rPPG \cite{ciftci2019fakecatcher}, blink patterns \cite{li2018blink} \\
\addlinespace
\midrule
Audio & ASVspoof 2019/2021 datasets and challenge reports \cite{asvspoof2019, asvspoof2021sum, asvspoof2021site}; DeepMind SynthID audio watermark \cite{deepmind2024synthid} \\
\addlinespace
\midrule
Provenance\newline /Policy & C2PA technical specification and overview \cite{c2paSite} \\
\bottomrule
\end{tabular}
\end{table}

\section*{Acknowledgments}
We thank the reviewers of the \textbf{MMAI Workshop on Multimodal AI}, held in conjunction with the \textbf{IEEE International Conference on Data Mining (ICDM) 2025}, for their constructive feedback. Their comments directly motivated the addition of a unified taxonomy, a threat matrix, temporal-forensics notes, retrieval deployment guidance, and a consolidated resource appendix.

\bibliographystyle{IEEEtrans}
\balance
\bibliography{ref}

\begin{thebibliography}{10}
\providecommand{\url}[1]{#1}
\csname url@samestyle\endcsname
\providecommand{\newblock}{\relax}
\providecommand{\bibinfo}[2]{#2}
\providecommand{\BIBentrySTDinterwordspacing}{\spaceskip=0pt\relax}
\providecommand{\BIBentryALTinterwordstretchfactor}{4}
\providecommand{\BIBentryALTinterwordspacing}{\spaceskip=\fontdimen2\font plus
\BIBentryALTinterwordstretchfactor\fontdimen3\font minus \fontdimen4\font\relax}
\providecommand{\BIBforeignlanguage}[2]{{%
\expandafter\ifx\csname l@#1\endcsname\relax
\typeout{** WARNING: IEEEtranS.bst: No hyphenation pattern has been}%
\typeout{** loaded for the language `#1'. Using the pattern for}%
\typeout{** the default language instead.}%
\else
\language=\csname l@#1\endcsname
\fi
#2}}
\providecommand{\BIBdecl}{\relax}
\BIBdecl

\bibitem{achiam2023gpt}
J.~Achiam, S.~Adler, S.~Agarwal, L.~Ahmad, I.~Akkaya, F.~L. Aleman, D.~Almeida, J.~Altenschmidt, S.~Altman, S.~Anadkat \emph{et~al.}, ``Gpt-4 technical report,'' \emph{arXiv preprint arXiv:2303.08774}, 2023.

\bibitem{agarwal2020phoneme}
S.~Agarwal and H.~Farid, ``Detecting deep-fake videos from phoneme--viseme mismatches,'' in \emph{Proceedings of the IEEE/CVF Conference on Computer Vision and Pattern Recognition Workshops (CVPRW)}, 2020.

\bibitem{asvspoof2021site}
{ASVspoof Challenge Organizers}, ``{ASVspoof} 2021 challenge website,'' \url{https://asvspoof.org/}, 2021, accessed 2025-09-26.

\bibitem{baevski2020wav2vec}
A.~Baevski, Y.~Zhou, A.~Mohamed, and M.~Auli, ``wav2vec 2.0: A framework for self-supervised learning of speech representations,'' \emph{Advances in neural information processing systems}, vol.~33, pp. 12\,449--12\,460, 2020.

\bibitem{bai2024ai}
J.~Bai, M.~Lin, G.~Cao, and Z.~Lou, ``{AI}-generated video detection via spatial-temporal anomaly learning,'' in \emph{Chinese Conference on Pattern Recognition and Computer Vision (PRCV)}, 2024.

\bibitem{bao2023fastdetectgpt}
F.~Bao, M.~Post, C.~Raffel, J.~He, Z.~He, and C.~Callison-Burch, ``Fast-detectgpt: Efficient zero-shot detection of machine-generated text via self-consistency,'' in \emph{The Twelfth International Conference on Learning Representations (ICLR)}, 2024.

\bibitem{bhattacharjee2024fighting}
A.~Bhattacharjee and H.~Liu, ``Fighting fire with fire: Can chatgpt detect ai-generated text?'' \emph{{ACM} SIGKDD Explorations Newsletter}, vol.~25, no.~2, pp. 14--21, 2024.

\bibitem{borji2023qualitative}
A.~Borji, ``Qualitative failures of image generation models and their application in detecting deepfakes,'' \emph{Image and Vision Computing}, vol. 136, p. 104771, 2023.

\bibitem{cao2025practical}
\BIBentryALTinterwordspacing
L.~Cao, \emph{A Practical Guide to Detect GenAI Content}, 1st~ed.\hskip 1em plus 0.5em minus 0.4em\relax Amazon, 2025, kindle Direct Publishing (KDP). [Online]. Available: \url{https://www.amazon.com/dp/B0F2ZKH2R4}
\BIBentrySTDinterwordspacing

\bibitem{chaka2024reviewing}
C.~Chaka, ``Reviewing the performance of ai detection tools in differentiating between ai-generated and human-written texts: A literature and integrative hybrid review,'' \emph{Journal of Applied Learning and Teaching}, vol.~7, no.~1, 2024.

\bibitem{chang2024what}
C.~Chang, Z.~Liu, X.~Lyu, and X.~Qi, ``What matters in detecting ai-generated videos like sora?'' in \emph{arXiv preprint arXiv:2406.19568}, 2024.

\bibitem{chen2023gpt}
Y.~Chen, H.~Kang, V.~Zhai, L.~Li, R.~Singh, and B.~Raj, ``{GPT}-sentinel: Distinguishing human and chatgpt generated content,'' \emph{arXiv preprint arXiv:2305.07969}, 2023.

\bibitem{ciftci2019fakecatcher}
U.~A. Ciftci, I.~Demir, and L.~Yin, ``{FakeCatcher}: Detection of synthetic portrait videos using biological signals,'' \emph{IEEE Transactions on Pattern Analysis and Machine Intelligence}, 2020, early version: arXiv:1901.02212 (2019).

\bibitem{c2paSite}
{Coalition for Content Provenance and Authenticity (C2PA)}, ``{C2PA} - verifying media content sources,'' \url{https://c2pa.org/}, 2025, accessed 2025-09-26.

\bibitem{comanducci2024fakemusiccaps}
L.~Comanducci, P.~Bestagini, and S.~Tubaro, ``{FakeMusicCaps}: A dataset for detection and attribution of synthetic music generated via text-to-music models,'' in \emph{arXiv preprint arXiv:2409.10684}, 2024.

\bibitem{crothers2023machine}
E.~N. Crothers, N.~Japkowicz, and H.~L. Viktor, ``Machine-generated text: A comprehensive survey of threat models and detection methods,'' \emph{IEEE Access}, vol.~11, pp. 70\,977--71\,002, 2023.

\bibitem{dolhansky2020deepfake}
B.~Dolhansky, J.~Bitton, B.~Pflaum, J.~Lu, R.~Howes, M.~Wang, and C.~C. Ferrer, ``The deepfake detection challenge (dfdc) dataset,'' \emph{arXiv preprint arXiv:2006.07397}, 2020.

\bibitem{aigcdBench}
{FDMAS Research Group}, ``Awesome {AIGC} detection benchmark (aigcdetect),'' \url{https://fdmas.github.io/AIGCDetect}, 2023, accessed 2025-09-26.

\bibitem{deepmind2024synthid}
{Google DeepMind}, ``Synthid for {AI}-generated audio,'' \url{https://deepmind.google/science/synthid/ai-generated-audio/}, 2024, accessed 2025-09-26.

\bibitem{aigcdPages}
{Gray Dove}, ``Awesome {AIGC} image detection,'' \url{https://github.com/graydove/Awesome-AIGC-Image-Detection}, 2023, accessed 2025-09-26.

\bibitem{hans2024spotting}
A.~Hans, A.~Schwarzschild, V.~Cherepanova, H.~Kazemi \emph{et~al.}, ``Spotting llms with binoculars: zero-shot detection of machine-generated text,'' in \emph{International Conference on Machine Learning (ICML)}, 2024.

\bibitem{ho2020denoising}
J.~Ho, A.~Jain, and P.~Abbeel, ``Denoising diffusion probabilistic models,'' \emph{Advances in Neural Information Processing Systems}, vol.~33, pp. 6840--6851, 2020.

\bibitem{hou2023semstamp}
L.~Hou, M.~Wang, X.~Liu, B.~Li, and Q.~Li, ``Semstamp: A semantic watermark for large language models,'' in \emph{Proceedings of the 2024 Conference of the North American Chapter of the Association for Computational Linguistics (NAACL)}, 2024.

\bibitem{hu2023radar}
X.~Hu, P.-Y. Chen, and T.~Y. Ho, ``{RADAR}: Robust ai-text detection via adversarial learning,'' in \emph{Advances in Neural Information Processing Systems}, 2023, pp. 15\,077--15\,095.

\bibitem{hussain2022forensic}
M.~Hussain \emph{et~al.}, ``Forensic audio authentication in law enforcement and court proceedings,'' \emph{ArXiv e-prints}, 2022, arXiv:2210.11273.

\bibitem{kamali2024distinguish}
N.~Kamali, K.~Nakamura, A.~Chatzimparmpas, J.~Hullman, and M.~Groh, ``How to distinguish ai-generated images from authentic photographs,'' \emph{arXiv preprint arXiv:2406.08651}, 2024.

\bibitem{karras2018progressive}
T.~Karras, T.~Aila, S.~Laine, and J.~Lehtinen, ``Progressive growing of {GAN}s for improved quality, stability, and variation,'' in \emph{International Conference on Learning Representations}, 2018.

\bibitem{kirchenbauer2023watermark}
J.~Kirchenbauer, J.~Geiping, Y.~Wen, J.~Katz, I.~Miers, and T.~Goldstein, ``A watermark for large language models,'' in \emph{Proceedings of the 40th International Conference on Machine Learning (ICML)}, ser. Proceedings of Machine Learning Research, vol. 202.\hskip 1em plus 0.5em minus 0.4em\relax PMLR, 2023.

\bibitem{kirchenbauer2023reliability}
J.~Kirchenbauer, J.~Geiping, Y.~Wen, M.~Shu, K.~Saifullah, K.~Kong, K.~Fernando, A.~Saha, M.~Goldblum, and T.~Goldstein, ``On the reliability of watermarks for large language models,'' \emph{arXiv preprint arXiv:2306.04634}, 2023.

\bibitem{krishna2024paraphrasing}
K.~Krishna, Y.~Song, M.~Karpinska, J.~Wieting, and M.~Iyyer, ``Paraphrasing evades detectors of ai-generated text, but retrieval is an effective defense,'' in \emph{NeurIPS}, 2024.

\bibitem{krishna2023paraphrasing}
K.~Krishna, V.~Zayats, S.~Agrawal, and M.~Iyyer, ``Paraphrasing evades detectors of {AI}-generated text,'' \emph{arXiv preprint arXiv:2303.13408}, 2023.

\bibitem{kuditipudi2023robust}
R.~Kuditipudi, J.~Thickstun, T.~Hashimoto, and P.~Liang, ``Robust distortion-free watermarks for language models,'' \emph{arXiv preprint arXiv:2307.15593}, 2023.

\bibitem{li2023origin}
L.~Li, P.~Wang, K.~Ren, T.~Sun, and X.~Qiu, ``Origin tracing and detecting of llms,'' \emph{arXiv preprint arXiv:2304.14072}, 2023.

\bibitem{li2024audio}
M.~Li, Y.~Ahmadiadli, and X.-P. Zhang, ``Audio anti-spoofing detection: A survey,'' \emph{arXiv preprint arXiv:2404.13914}, 2024.

\bibitem{li2024detectingmachine}
Y.~Li, H.~Li, L.~Specia, and B.~W. Schuller, ``{M6}: Multi-generator, multi-domain, multi-lingual and cultural, multi-genres, multi-instrument machine-generated music detection databases,'' \emph{arXiv preprint arXiv:2412.06001}, 2024.

\bibitem{li2024milling}
Y.~Li, M.~Milling, L.~Specia, and B.~W. Schuller, ``to ai-generated music detection: A pathway and overview,'' \emph{arXiv preprint arXiv:2412.00571}, 2024.

\bibitem{li2024aigmdetection}
Y.~Li, Q.~Sun, H.~Li, L.~Specia, and B.~W. Schuller, ``Detecting machine-generated music with explainability: A challenge and early benchmarks,'' \emph{arXiv preprint arXiv:2412.13421}, 2024.

\bibitem{li2018blink}
Y.~Li, M.-C. Chang, and S.~Lyu, ``In ictu oculi: Exposing ai-created fake videos by detecting eye blinking,'' in \emph{2018 IEEE International Workshop on Information Forensics and Security (WIFS)}, 2018, pp. 1--7.

\bibitem{lin2024detecting}
L.~Lin, N.~Gupta, Y.~Zhang, H.~Ren, C.~H. Liu, F.~Ding, X.~Wang, X.~Li, L.~Verdoliva, and S.~Hu, ``Detecting multimedia generated by large {AI} models: A survey,'' \emph{arXiv preprint arXiv:2402.00045}, 2024.

\bibitem{liu2024semantic}
A.~Liu, L.~Pan, X.~Hu, S.~Meng, and L.~Wen, ``A semantic invariant robust watermark for large language models,'' \emph{ICLR}, 2024.

\bibitem{liu2019roberta}
Y.~Liu, M.~Ott, N.~Goyal, J.~Du, M.~Joshi, D.~Chen, O.~Levy, M.~Lewis, L.~Zettlemoyer, and V.~Stoyanov, ``{RoBERTa}: A robustly optimized {BERT} pretraining approach,'' \emph{arXiv preprint arXiv:1907.11692}, 2019.

\bibitem{mavali2024fake}
S.~Mavali, J.~Ricker, D.~Pape, Y.~Sharma, A.~Fischer, and L.~Sch{\"o}nherr, ``Fake it until you break it: on the adversarial robustness of ai-generated image detectors,'' \emph{arXiv preprint arXiv:2410.01574}, 2024.

\bibitem{mireshghallah2024smaller}
N.~S. Mireshghallah, J.~Mattern, S.~Gao, R.~Shokri, and T.~Berg-Kirkpatrick, ``Smaller language models are better zero-shot machine-generated text detectors,'' \emph{Proceedings of the EACL}, pp. 278--293, 2024.

\bibitem{mitchell2023detectgpt}
E.~Mitchell, Y.~Lee, A.~Khazatsky, C.~D. Manning, and C.~Finn, ``Detectgpt: Zero-shot machine-generated text detection using probability curvature,'' in \emph{Proceedings of the 40th International Conference on Machine Learning (ICML)}, ser. Proceedings of Machine Learning Research, vol. 202.\hskip 1em plus 0.5em minus 0.4em\relax PMLR, 2023.

\bibitem{muller2022does}
N.~M. M{\"u}ller, P.~Czempin, F.~Dieckmann, A.~Froghyar, and K.~B{\"o}ttinger, ``Does audio deepfake detection generalize?'' in \emph{INTERSPEECH 2022}, 2022.

\bibitem{nguyen2024videofact}
T.~D. Nguyen, S.~Fang, and M.~C. Stamm, ``{Videofact}: Detecting video forgeries using attention, scene context, and forensic traces,'' in \emph{WACV}, 2024, pp. 8563--8573.

\bibitem{nyberg2022bard}
E.~P. Nyberg, A.~E. Nicholson, K.~B. Korb, M.~Wybrow, I.~Zukerman, S.~Mascaro, S.~Thakur, A.~Oshni~Alvandi, J.~Riley, R.~Pearson \emph{et~al.}, ``Bard: A structured technique for group elicitation of bayesian networks to support analytic reasoning,'' \emph{Risk Analysis}, vol.~42, no.~6, pp. 1155--1178, 2022.

\bibitem{ong2023applying}
I.~Ong and B.~K. Quek, ``Applying ensemble methods to model-agnostic machine-generated text detection,'' \emph{arXiv preprint arXiv:2406.12570}, 2023.

\bibitem{pascu2024generalizable}
O.~Pascu, A.~Stan, D.~Oneata, E.~Oneata, and H.~Cucu, ``Towards generalisable and calibrated audio deepfake detection with self-supervised representations,'' in \emph{Interspeech}, 2024, pp. 4828--4832.

\bibitem{radford2019language}
A.~Radford, J.~Wu, R.~Child, D.~Luan, D.~Amodei, I.~Sutskever \emph{et~al.}, ``Language models are unsupervised multitask learners,'' \emph{OpenAI blog}, vol.~1, no.~8, p.~9, 2019.

\bibitem{rao2022melody}
K.~S. Rao and P.~P. Das, ``Melody extraction from polyphonic music by deep learning approaches: a review,'' \emph{arXiv preprint arXiv:2202.01078}, 2022.

\bibitem{pruthi2024revisiting}
S.~Rastogi and D.~Pruthi, ``Revisiting the robustness of watermarking to paraphrasing attacks,'' in \emph{Proceedings of the 2024 Conference on Empirical Methods in Natural Language Processing (EMNLP)}.\hskip 1em plus 0.5em minus 0.4em\relax Miami, Florida, USA: Association for Computational Linguistics, 2024, pp. 18\,100--18\,110.

\bibitem{ravanelli2018speaker}
M.~Ravanelli and Y.~Bengio, ``Speaker recognition from raw waveform with sincnet,'' in \emph{2018 IEEE Spoken Language Technology Workshop (SLT)}.\hskip 1em plus 0.5em minus 0.4em\relax IEEE, 2018, pp. 1021--1028.

\bibitem{rossler2019faceforensics++}
A.~Rossler, D.~Cozzolino, L.~Verdoliva, C.~Riess, J.~Thies, and M.~Nie{\ss}ner, ``Faceforensics++: Learning to detect manipulated facial images,'' in \emph{Proceedings of the IEEE/CVF international conference on computer vision}, 2019, pp. 1--11.

\bibitem{saberi2024robustness}
M.~Saberi, V.~Sadasivan, K.~Rezaei, A.~Kumar, A.~Chegini, W.~Wang, and S.~Feizi, ``Robustness of ai-image detectors: fundamental limits and practical attacks,'' \emph{ICLR}, 2024.

\bibitem{sadasivan2023can}
V.~S. Sadasivan, A.~Kumar, S.~Balasubramanian, W.~Wang, and S.~Feizi, ``Can {AI}-generated text be reliably detected?'' \emph{arXiv preprint arXiv:2303.11156}, 2023.

\bibitem{saskoro2024detection}
R.~A.~F. Saskoro, N.~Yudistira, and T.~N. Fatyanosa, ``Detection of ai-generated images from various generators using gated expert convolutional neural network,'' \emph{IEEE Access}, 2024.

\bibitem{schuhmann2022laion}
C.~Schuhmann, R.~Beaumont, R.~Vencu, C.~Gordon, R.~Wightman, M.~Cherti, T.~Coombes, A.~Katta, C.~Mullis, M.~Wortsman \emph{et~al.}, ``Laion-5b: An open large-scale dataset for training next generation image-text models,'' \emph{Advances in Neural Information Processing Systems}, vol.~35, pp. 25\,278--25\,294, 2022.

\bibitem{sharma2024review}
S.~Sharma, J.~J. Zou, G.~Fang, P.~Shukla, and W.~Cai, ``A review of image watermarking for identity protection and verification,'' \emph{Multimedia Tools and Applications}, vol.~83, no.~11, pp. 31\,829--31\,891, 2024.

\bibitem{su2023detectllm}
J.~Su, T.~Zhuo, D.~Wang, and P.~Nakov, ``{DetectLLM}: Leveraging log rank information for zero-shot detection of machine-generated text,'' \emph{Findings of the Association for Computational Linguistics: EMNLP}, pp. 12\,395--12\,412, 2023.

\bibitem{taguchi2024impact}
K.~Taguchi, Y.~Gu, and K.~Sakurai, ``The impact of prompts on zero-shot detection of ai-generated text,'' \emph{arXiv preprint arXiv:2403.20127}, 2024.

\bibitem{tang2024science}
R.~Tang, Y.-N. Chuang, and X.~Hu, ``The science of detecting llm-generated text,'' \emph{Communications of the ACM}, vol.~67, no.~4, pp. 50--59, 2024.

\bibitem{tian2024multiscale}
Y.~Tian, H.~Chen, X.~Wang \emph{et~al.}, ``Multiscale positive-unlabeled detection of ai-generated texts,'' in \emph{International Conference on Learning Representations (ICLR)}, 2024.

\bibitem{tiwari2024ai}
A.~K. Tiwari, A.~Sharma, P.~Rayakar, and M.~K. Bhavriya, ``{AI}-generated video forgery detection and authentication,'' in \emph{IEEE I2CT}, 2024, pp. 1--8.

\bibitem{vahdati2024beyond}
D.~S. Vahdati, T.~D. Nguyen, H.~Azizpour, and M.~C. Stamm, ``Beyond deepfake images: detecting ai-generated videos,'' in \emph{CVPR}, 2024.

\bibitem{valiaiev2024detection}
D.~Valiaiev, ``Detection of machine-generated text: Literature survey,'' \emph{arXiv preprint arXiv:2402.01642}, 2024.

\bibitem{asvspoof2019}
X.~Wang, J.~Yamagishi, A.~Nautsch, N.~Evans, T.~Kinnunen, M.~Todisco, H.~Delgado, M.~Sahidullah, V.~Vestman, K.~A. Lee \emph{et~al.}, ``{ASVspoof} 2019: Automatic speaker verification spoofing and countermeasures challenge,'' \emph{arXiv preprint arXiv:1911.01601}, 2019.

\bibitem{wang2020asvspoof}
X.~Wang, J.~Yamagishi, M.~Todisco, H.~Delgado, A.~Nautsch, N.~Evans, M.~Sahidullah, V.~Vestman, T.~Kinnunen, K.-A. Lee \emph{et~al.}, ``{ASVspoof} 2019: A large-scale public database of synthesized, converted and replayed speech,'' in \emph{Computer Speech \& Language}, vol.~64.\hskip 1em plus 0.5em minus 0.4em\relax Elsevier, 2020, p. 101114.

\bibitem{wu2024survey}
J.~Wu, S.~Yang, R.~Zhan, Y.~Yuan, D.~F. Wong, and L.~S. Chao, ``A survey on llm-gernerated text detection: Necessity, methods, and future directions,'' \emph{arXiv preprint arXiv:2310.14724}, 2024.

\bibitem{asvspoof2021sum}
J.~Yamagishi, N.~Evans, M.~Todisco, H.~Delgado, X.~Wang, T.~Kinnunen \emph{et~al.}, ``{ASVspoof} 2021: Accelerating progress in spoofed and deepfake speech detection,'' in \emph{Proceedings of the ASVspoof 2021 Workshop}, 2021.

\bibitem{yan2024sanity}
S.~Yan, O.~Li, J.~Cai, Y.~Hao, X.~Jiang, Y.~Hu, and W.~Xie, ``A sanity check for {AI}-generated image detection,'' \emph{arXiv preprint arXiv:2406.19435}, 2024.

\bibitem{yang2023survey}
X.~Yang, L.~Pan, X.~Zhao, H.~Chen, L.~Petzold, W.~Y. Wang, and W.~Cheng, ``A survey on detection of llms-generated content,'' \emph{arXiv preprint arXiv:2310.15654}, 2023.

\bibitem{yi2023audio}
J.~Yi, C.~Wang, J.~Tao, X.~Zhang, C.~Zhang, and Y.~Zhao, ``Audio deepfake detection: A survey,'' \emph{arXiv preprint arXiv:2308.14970}, 2023.

\bibitem{zeng2024detecting}
Z.~Zeng, S.~Liu, L.~Sha, Z.~Li, K.~Yang, S.~Liu, and G.~Chen, ``Detecting ai-generated sentences in realistic human-ai collaborative hybrid texts: Challenges, strategies, and insights,'' in \emph{arXiv preprint arXiv:2403.03506}, 2024.

\bibitem{zhao2024provable}
X.~Zhao, P.~Ananth, L.~Li, and Y.-X. Wang, ``Provable robust watermarking for {AI}-generated text,'' in \emph{The Twelfth International Conference on Learning Representations (ICLR)}, 2024.

\bibitem{zhu2023genimage}
M.~Zhu, H.~Chen, Q.~Yan, X.~Huang, G.~Lin, W.~Li, Z.~Tu, H.~Hu, J.~Hu, and Y.~Wang, ``{GenImage}: A million-scale benchmark for detecting {AI}-generated image,'' \emph{arXiv preprint arXiv:2306.08571}, 2023.

\bibitem{genimageNeurIPS23}
------, ``{GenImage}: A million-scale benchmark for detecting {AI}-generated image,'' in \emph{Proceedings of the 37th Conference on Neural Information Processing Systems (NeurIPS), Track on Datasets and Benchmarks}, 2023.

\end{thebibliography}

\end{document}